\documentclass[sigconf,natbib=true,anonymous=false]{acmart}
\AtBeginDocument{%
  }

\setcopyright{acmlicensed}
\copyrightyear{2026}
\acmYear{2026}
\acmDOI{XXXXXXX.XXXXXXX}
\acmConference[SIGIR '26]{the 49th International ACM SIGIR Conference on Research and Development in Information Retrieval}{July}{Melbourne, Australia}
\acmISBN{978-1-4503-XXXX-X/2026/06}

\usepackage{amsmath}
\usepackage{amsfonts}

\usepackage{amssymb}
\usepackage{graphicx}
\usepackage{enumitem}

\usepackage[dvipsnames,svgnames]{xcolor}
\usepackage{latexsym}
\usepackage{multirow} 
\usepackage[most]{tcolorbox}
 \usepackage{xcolor}
 \usepackage{subcaption}
\usepackage{xcolor,pifont}
\newcommand{\xmark}{\ding{55}}%
\newcommand{\cmark}{\ding{52}}%
\newcommand*\colourcheck[1]{%
  \expandafter\newcommand\csname #1check\endcsname{\textcolor{#1}{\ding{52}}}%
}
\colourcheck{blue}
\colourcheck{green}
\colourcheck{red}
\usepackage{adjustbox}
\usepackage{booktabs}
\usepackage{multirow}
\usepackage{array}
\usepackage{makecell}
\usepackage{array}
\newcolumntype{C}[1]{>{\centering\arraybackslash}p{#1}}

\usepackage{tabularx}
\usepackage{placeins}
\setlist[itemize]{leftmargin=*}
\setlist[enumerate]{leftmargin=*}
\begin{document}

\title{Diagnosing Causal Reasoning in Vision-Language Models via Structured Relevance Graphs}


\author{Dhita Putri Pratama}
\affiliation{%
  \institution{University of Melbourne}
  \city{Melbourne}
  \country{Australia}}

\author{Soyeon Caren Han}
\affiliation{%
  \institution{University of Melbourne}
  \city{Melbourne}
  \country{Australia}
}

\author{Yihao Ding}
\affiliation{%
 \institution{The University of Western Australia}
 \city{Perth}
 \country{Australia}}


\begin{abstract}
Large Vision-Language Models (LVLMs) achieve strong performance on visual question answering benchmarks, yet often rely on spurious correlations rather than genuine causal reasoning. Existing evaluations primarily assess the correctness of the answers, making it unclear whether failures arise from limited reasoning capability or from misidentifying causally relevant information.
We introduce \textbf{Vision-Language Causal Graphs (VLCGs)}, a structured, query-conditioned representation that explicitly encodes causally relevant objects, attributes, relations, and scene-grounded assumptions. Building on this representation, we present \textbf{ViLCaR}, a diagnostic benchmark comprising tasks for Causal Attribution, Causal Inference, and Question Answering, along with graph-aligned evaluation metrics that assess relevance identification beyond final answer accuracy.
Experiments in state-of-the-art LVLMs show that injecting structured relevance information significantly improves attribution and inference consistency compared to zero-shot and standard in-context learning. These findings suggest that current limitations in LVLM causal reasoning stem primarily from insufficient structural guidance rather than a lack of reasoning capacity.
\end{abstract}



\maketitle

\begin{table}[t]
\centering
\renewcommand{\arraystretch}{1.1}

\begin{adjustbox}{max width=\linewidth}
\begin{tabular}{
    p{2.3cm}
    p{0.8cm}
    >{\centering\arraybackslash}p{0.55cm}
    >{\centering\arraybackslash}p{0.55cm}
    >{\centering\arraybackslash}p{0.7cm}
    >{\centering\arraybackslash}p{0.55cm}
    >{\centering\arraybackslash}p{0.55cm}
    >{\centering\arraybackslash}p{0.55cm}
    >{\centering\arraybackslash}p{0.55cm}
}
\toprule

\multirow{2}{*}{\small Datasets} 
& \multirow{2}{*}{\small Size}
& \multicolumn{4}{c}{\small Graph Component}
& \multicolumn{3}{c}{\small Tasks} \\
\cmidrule(lr){3-6} \cmidrule(lr){7-9}
& 
& \small OA 
& \small OR 
& \small CAss 
& \small CD 
& \small CA 
& \small CI 
& \small QA \\

\midrule

\small VQA \shortcite{vqa_2015, vqa_v2_2017} 
& \small 1.1M  
& \xmark & \xmark & \xmark & \xmark 
& \xmark & \xmark & \cmark \\

\small Visual7W \shortcite{visual7w_2016} 
& \small 328K 
& \xmark & \xmark & \xmark & \xmark 
& \xmark & \xmark & \cmark \\

\small V-Genome \shortcite{visual_genome_2017} 
& \small 1.7M 
& \cmark & \cmark & \xmark & \xmark 
& \xmark & \xmark & \cmark \\

\small VCR \shortcite{vcr_2019} 
& \small 290K 
& \xmark & \xmark & \xmark & \xmark 
& \xmark & \xmark & \cmark \\

\small OK-VQA \shortcite{okvqa} 
& \small 14k 
& \xmark & \xmark & \xmark & \xmark 
& \xmark & \xmark & \cmark \\

\small CoSIm \shortcite{cosim_2022} 
& \small 3.5K 
& \xmark & \xmark & \xmark & \xmark 
& \xmark & \xmark & \cmark \\

\small CELLO \shortcite{cello_2024} 
& \small 14K 
& \xmark & \cmark & \xmark & \xmark 
& \cmark & \xmark & \cmark \\

\midrule

\small \textbf{ViLCaR (Ours)} 
& \small 12.5K 
& \greencheck & \greencheck & \greencheck & \greencheck 
& \greencheck & \greencheck & \greencheck \\

\bottomrule

\end{tabular}
\end{adjustbox}

\caption{
Comparison of the ViLCaR with existing works that include causal reasoning tasks. 
\textbf{OA}, \textbf{OR}, \textbf{CAss}, and \textbf{CD} denote \textit{Object Attribute}, \textit{Object Relation}, \textit{Causal Assumption}, and \textit{Context Dependence}. 
\textbf{CA}, \textbf{CI}, and \textbf{QA} represent \textit{Causal Attribution}, \textit{Causal Inference}, and \textit{Question Answering}.
}
\label{tab:data_comparison}
\vspace{-2.2em}
\end{table}

\section{Introduction}

Large Vision-Language Models (LVLMs) have demonstrated strong performance on visual question answering and multimodal reasoning benchmarks. However, high answer accuracy does not necessarily imply faithful or causally grounded reasoning. Models may produce correct answers while relying on spurious visual cues or superficial associations, resulting in explanations that are inconsistent or unfaithful \cite{making_reasoning_matter, unfaithful_reasoning_in_cot}. This discrepancy reveals a key limitation of current evaluation paradigms: they primarily measure \emph{prediction correctness}, but do not diagnose whether models correctly identify the causally relevant information required for valid reasoning.
We argue that many failures in visual causal reasoning stem from errors in \emph{relevance identification}. Before performing inference, a model must determine which objects, attributes, relations, and contextual assumptions are causally relevant to a query. When this identification stage fails, downstream reasoning becomes brittle or misleading, even if the final answer is correct. Despite its importance, this step remains largely unexamined in existing benchmarks.
As summarized in Table~\ref{tab:data_comparison}, most VQA-style datasets \cite{vqa_2015, vqa_v2_2017, visual7w_2016, vcr_2019, okvqa, cosim_2022} lack explicit causal structure, providing neither attribute-level dependencies nor scene-grounded assumptions. Although CELLO \cite{cello_2024} introduces object-level causal graphs, it does not model attribute-level factors or contextual assumptions necessary for fine-grained causal attribution and counterfactual reasoning. Moreover, evaluation protocols predominantly rely on answer accuracy, limiting the ability to distinguish perception errors from flawed causal reasoning.
To address these gaps, we introduce \textbf{Vision-Language Causal Graphs (VLCGs)}, a structured, query-conditioned representation that explicitly encodes objects, their attributes, inter-object relations, and scene-grounded assumptions as directed causal dependencies. Unlike standard scene graphs \cite{visual_genome_2017} or abstract causal models \cite{causality_pearl_2009}, VLCGs are designed to capture \emph{question-driven causal relevance} in visual contexts.
Building on this representation, we present \textbf{ViLCaR}, a diagnostic benchmark comprising three tasks: (1) \emph{Causal Attribution (CA)}, which evaluates whether models correctly identify causally relevant attributes; (2) \emph{Causal Inference (CI)}, which assesses the consistency of reasoning chains grounded in attributes and assumptions; and (3) \emph{Question Answering (QA)}, which measures final prediction accuracy. We further introduce graph-aligned evaluation metrics that disentangle relevance identification from answer correctness.
Our experiments show that simply providing question--answer exemplars does not reliably improve causal reasoning. In contrast, prompting with structured causal graphs significantly improves attribution and inference consistency. These findings suggest that current limitations in LVLM causal reasoning arise less from an inherent inability to reason and more from insufficient structural guidance. Main contributions are summarized:
\begin{itemize}
\item We propose VLCGs, a structured representation for modeling query-conditioned causal relevance in multimodal reasoning.
\item We introduce ViLCaR, a diagnostic benchmark enabling fine-grained analysis of causal attribution and inference.
\item We present graph-aligned evaluation metrics that disentangle relevance identification from final answer accuracy.
\end{itemize}

\section{VilCaR}
\subsection{Vision-Language Causal Graphs (VLCGs)}

We introduce \textbf{VLCGs}, a structured representation designed to model \emph{question conditioned causal relevance} in visual reasoning. 
Formally, given an image $I$ and a question $q$, a VLCG is a directed graph 
$G = (V, E, A)$ where:
(i) $V$ denotes scene-grounded entities (objects and abstract concepts),
(ii) $E$ denotes directed dependencies encoding attributes and inter-object relations, and 
(iii) $A$ denotes a set of explicit causal assumptions required to justify the correct answer.
Unlike traditional scene graphs that describe perceptual structure, VLCGs explicitly encode \emph{causally relevant} elements with respect to a specific image--question pair. The graph therefore represents not the full scene, but the minimal set of objects, attributes, relations, and contextual assumptions necessary to support a plausible causal explanation.
Crucially, VLCGs incorporate scene-grounded \emph{assumptions} following the structural view of causality \cite{causality_pearl_2009}. These assumptions encode implicit cultural or contextual knowledge that cannot be directly inferred from visual perception alone (e.g., attire as an indicator of a wedding ceremony). By integrating observable attributes with explicit assumptions, VLCGs capture the mechanism linking visual evidence to causal conclusions.

\begin{figure}[t]
  \includegraphics[width=0.98\columnwidth]{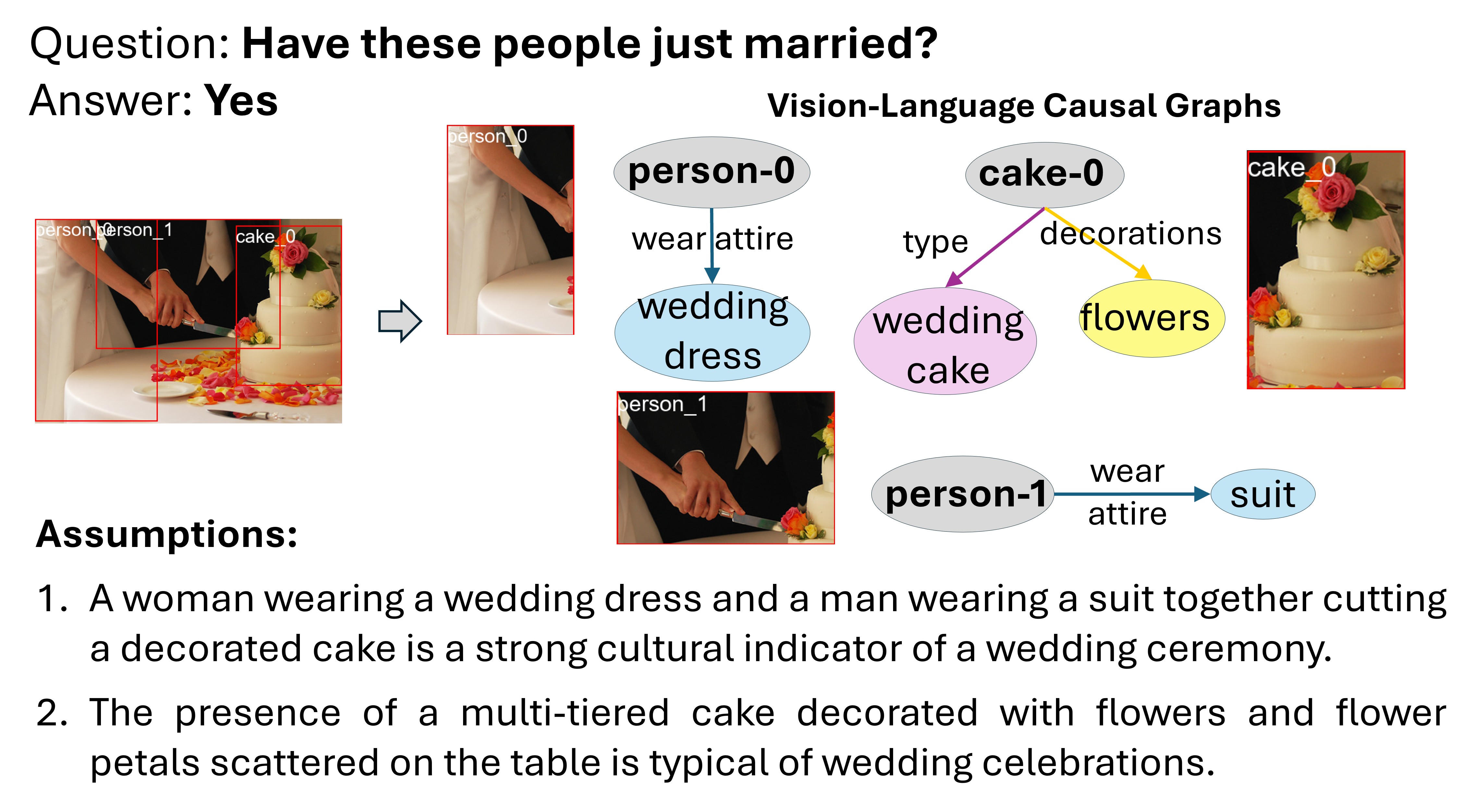 }
  \caption{Example of a VLCG. Given an image-question pair (“Have these people just married?”), the graph encodes causally relevant objects (e.g., persons, cake), attributes (wedding dress, suit), relations (wear), and scene-grounded assumptions linking visual evidence to the conclusion. Unlike scene graphs, VLCGs capture question-conditioned causal relevance rather than complete perceptual structure.}
  \label{fig:vilcar_format}
  \vspace{-1.5em}
\end{figure} 

\subsection{ViLCaR Tasks}
Building on VLCGs, we construct \textbf{ViLCaR}, a diagnostic benchmark for evaluating visual causal reasoning along three complementary dimensions: CA, CI, and QA.
\begin{enumerate}
    \item\textbf{Causal Attribution (CA).} Given $(I, q)$, the model must identify the set of causally relevant attributes or variables that influence the answer. This task evaluates whether the model selects appropriate causal factors rather than relying on spurious cues.
    
    \item\textbf{Causal Inference (CI).} The model must generate a reasoning chain that plausibly connects identified attributes and assumptions to the final answer. Reasoning is considered \emph{plausible} when the synthesized causal factors support a coherent conclusion.
    
    \item\textbf{Question Answering (QA).} The model predicts the final answer to $(I, q)$. This task measures outcome accuracy but does not, by itself, guarantee correct relevance identification or inferential validity.
\end{enumerate}

\begin{figure}[t]
  \includegraphics[width=0.99\columnwidth]{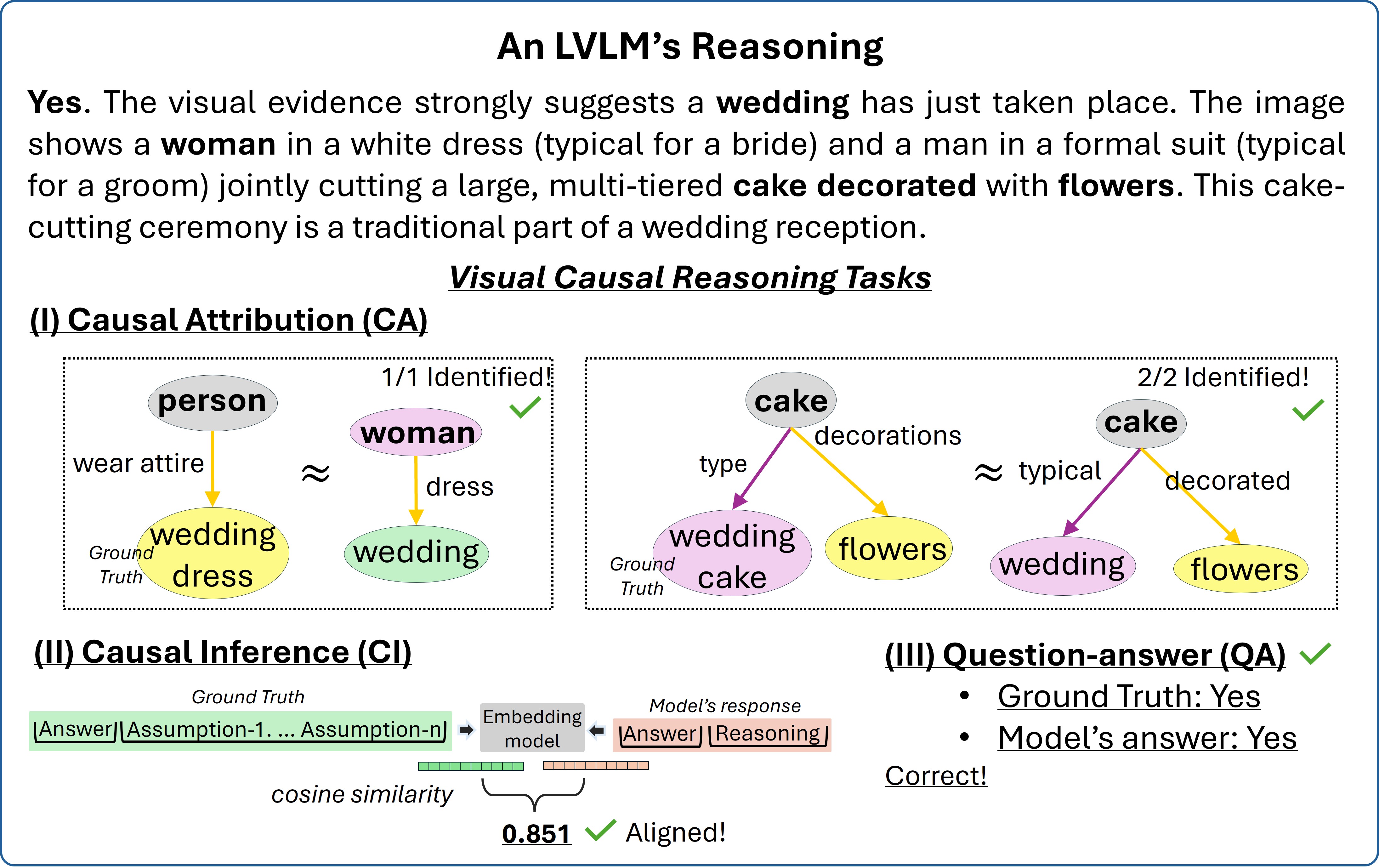}
  \caption{Three diagnostic tasks in ViLCaR derived from the verified and pruned VLCGs: CA, CI, and QA.}
  \label{fig:vilcar_task}
\end{figure}

\noindent These tasks disentangle three stages of visual causal reasoning: identifying relevant variables, composing a valid causal mechanism, and producing the final prediction. The overall pipeline is illustrated conceptually in Figure~\ref{fig:vilcar_task}.

\section{Dataset Construction}
In this paper, we construct ViLCaR from existing visual question answering datasets, primarily VQA \cite{vqa_2015, vqa_v2_2017} and VCR \cite{vcr_2019}, and transform them into structured and question-conditioned causal reasoning instances. 

\noindent\textbf{1) Question Selection and Causal Filtering.}
We first filter out questions that can be solved via object detection, spatial lookup, opinion-based judgment, or low-level perceptual cues. Remaining questions are categorized according to the Ladder of Causation \cite{book_of_why_pearl_2018}, retaining instances that require associative, interventional, or counterfactual reasoning. This step ensures that each instance requires non-trivial causal grounding.

\noindent\textbf{2) Graph Generation.}
For each image--question pair $(I, q)$, we construct a preliminary Vision-Language Causal Graph (VLCG) using LVLM-based prompting. The model is instructed to extract:
(i) causally relevant objects,
(ii) their attributes,
(iii) inter-object relations, and
(iv) explicit assumptions necessary to justify the answer.
The resulting graph encodes candidate causal factors rather than a full scene description.

\noindent\textbf{3) Graph Verification and Grounding.}
To reduce hallucinated entities and attributes, we validate graph components using independent vision models. Object nodes are aligned with detections from closed-set \cite{codetr_2023} and open-vocabulary detectors \cite{grounding_dino_2025, yolo_world_2024}, while attributes and relations are validated via image–text similarity scoring using CLIP-Score \cite{clip_score_2021}. Elements that fail grounding thresholds are removed. This step ensures that retained graph components are visually supported.

\noindent\textbf{4) Minimal Causal Pruning.}
To obtain a \emph{minimal sufficient} causal graph, we iteratively remove nodes and edges that are not required to derive the correct answer. An LLM, given only the graph (without the image), evaluates whether the remaining structure is sufficient to answer the question. Components that do not affect answer validity are pruned. The final VLCG therefore represents the smallest set of causally relevant elements supporting the ground-truth answer.

\noindent\textbf{5) Quality Control.}
We conduct human validation on 30 randomly sampled VLCGs with 15 annotators. Each graph is evaluated according to four criteria: correctness, relevance, sufficiency, and assumption strength, at the levels of objects, attributes, and relations. 
Disagreement rates remain below 15\% across all criteria, indicating consistent annotation quality. 
Notably, the near-equal distribution between ``agree'' and ``strongly agree'' responses highlights the inherent subjectivity of visual causal reasoning, even among humans.

\section{Data Statistics}
\begin{figure}[!t]
  \centering
  \begin{minipage}[t]{.55\linewidth}
    \subcaptionbox{Top-5 Objects}
      {\includegraphics[width=0.89\linewidth, height=4cm]{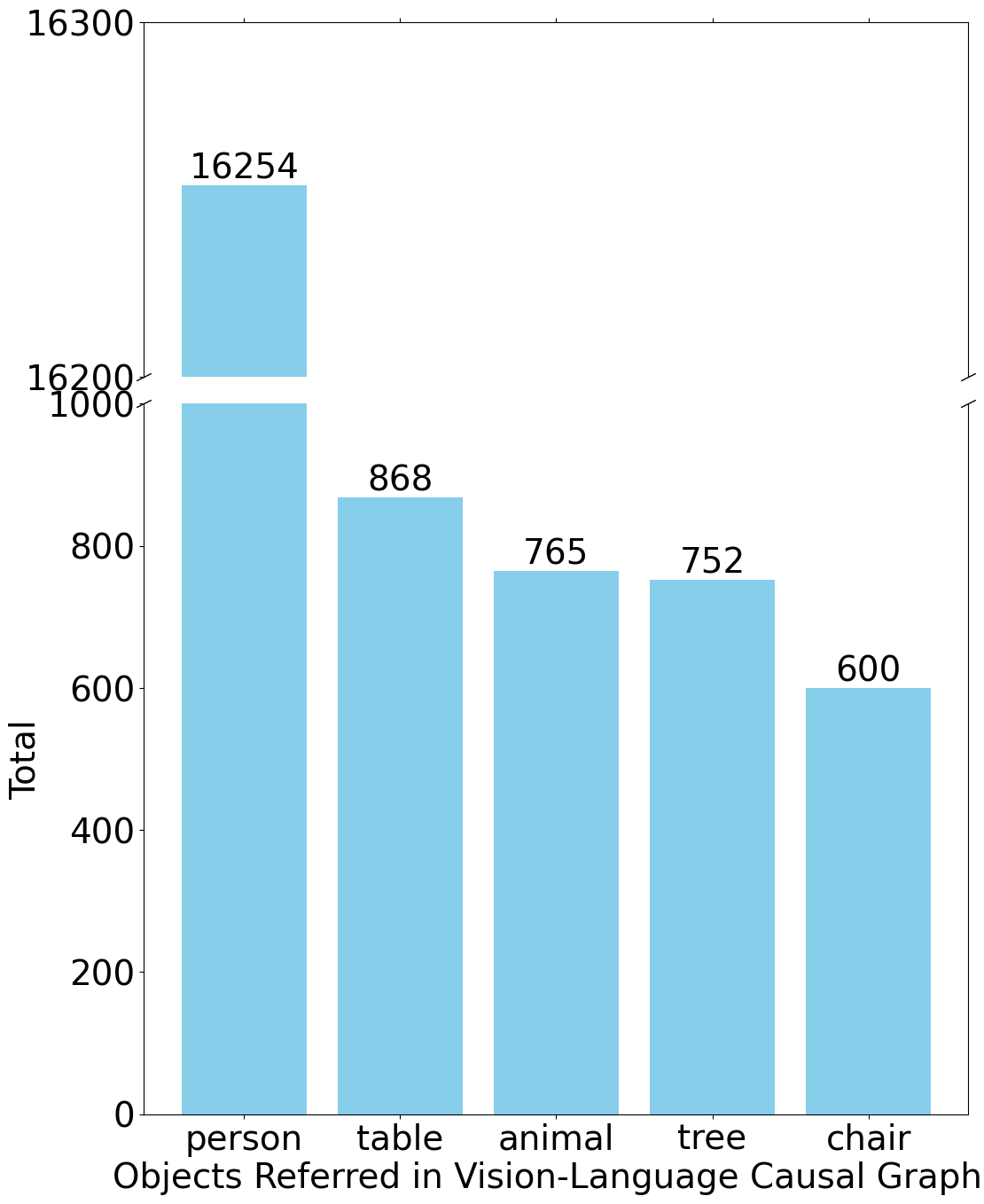}}%
  \end{minipage}%
  \hfill
  \begin{minipage}[b]{.42\linewidth}
    \subcaptionbox{Attribute: Word Cloud}
      {\includegraphics[width=0.9\linewidth, height=1.7cm]{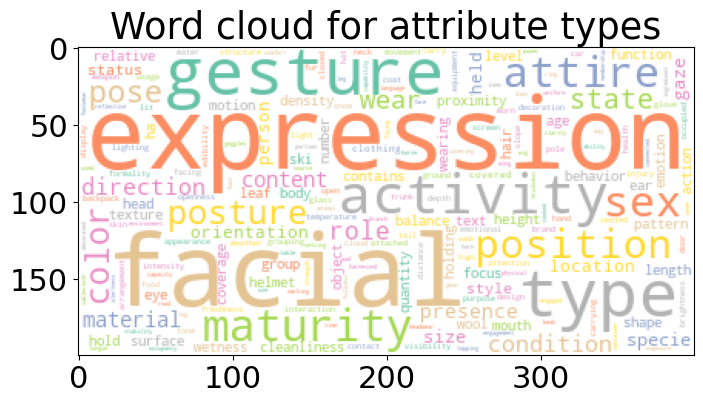}}
    \subcaptionbox{Relation: Word Cloud}
      {\includegraphics[width=0.9\linewidth, height=1.7cm]{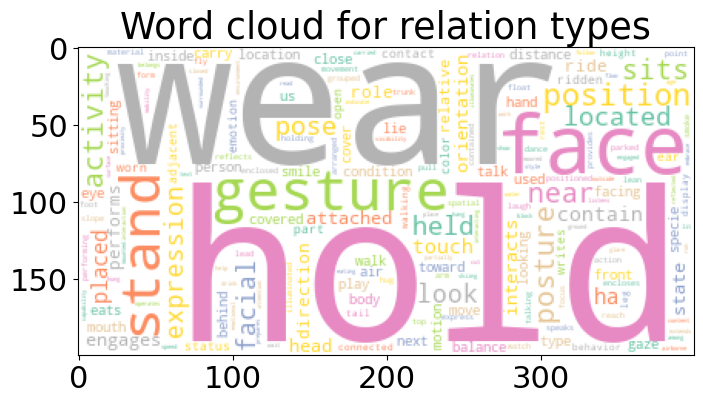}}%
  \end{minipage}%
  \caption
    {%
      A brief data statistics of VLCGs, with `\textbf{person}', mental states (e.g., facial expression), and physical relationships of the object (e.g., hold) being the most frequent objects, object characteristics, object relations, respectively.%
      \label{fig:vlcg_complexity}%
    }%
\end{figure}

We analyze the structural properties of the constructed VLCGs in Figure~\ref{fig:vlcg_complexity}. 
As shown in Figure~\ref{fig:vlcg_complexity}(a), \textit{person} is the most frequently referenced object, reflecting the prevalence of human-centered causal reasoning scenarios in ViLCaR. 
The attribute distribution in Figure~\ref{fig:vlcg_complexity}(b) is dominated by mental-state and role-related descriptors (e.g., facial expression, gesture), indicating that many causal inferences rely on socially grounded or affective cues rather than purely physical attributes. 
Similarly, the relation distribution in Figure~\ref{fig:vlcg_complexity}(c) highlights common physical interactions such as \textit{hold} and \textit{wear}, suggesting that causal reasoning in ViLCaR frequently integrates both social semantics and observable object interactions. 
Overall, these statistics confirm that VLCGs capture a mixture of perceptual evidence and higher-level contextual attributes, aligning with our goal of modeling question-conditioned causal relevance rather than purely spatial structure.

\section{Experiment Setup}
We evaluate whether structured VLCGs improve visual causal reasoning beyond standard prompting strategies. The main question is whether injecting explicit causal relevance signals enhances attribution and inference, rather than merely improving answer accuracy.

We benchmark Qwen2.5-VL-7B on ViLCaR using an 80/10/10 train--validation--test split. All reported results are computed on the test set. We compare three settings:
\begin{enumerate}
    \item\textbf{Zero-shot.} The model receives only the image and question and generates both reasoning and answer.
    \item\textbf{Standard ICL.} The model is provided with question-answer exemplars without structured graphs, testing whether unstructured few-shot prompting improves reasoning.  
    \item\textbf{VLCG-Augmented Prompting.} We inject structured causal graphs into the prompt and report results from the best-performing configuration.
\end{enumerate}

\noindent We evaluate performance along three diagnostic dimensions. By reporting CA, CI, and QA jointly, we disentangle correctness from causal reasoning quality.

\noindent\textbf{(a) Causal Attribution (CA).} We measure whether the model correctly identifies causally relevant attributes present in the VLCG. Given open-vocabulary reasoning outputs, we align reasoning tokens with graph elements using a semantic similarity score:
\begin{equation}
\label{eq:sim_score_combination}
S = S_{BERT} \times S_{W2V} \times \text{(}1 \text{ + } P_{unigram}\text{)}
\end{equation}
where $S_{BERT}$ and $S_{W2V}$ denote contextual and distributional similarity, and $P_{unigram}$ captures lexical overlap and serves as the weight importance for any token overlaps. A triplet (e.g., <cake, typical, wedding> is considered correctly identified if the similarity of at least 50\% triplet elements exceed a fixed threshold.

\noindent\textbf{Causal Inference (CI).}
CI evaluates whether the identified attributes are coherently composed into a valid reasoning chain that supports the final prediction.
Specifically, we use an LLM-based alignment protocol where a separate evaluator model is prompted to compare:
(i) the generated reasoning, and  
(ii) the gold causal assumptions encoded in the VLCG.
The evaluator assesses whether the reasoning:
(a) links relevant attributes to the outcome,
(b) maintains logical consistency, and
(c) avoids unsupported assumptions.
The CI score reflects agreement between generated reasoning and the structured causal path, averaged across test instances.

\noindent\textbf{(c) Question Answering (QA).}
QA reports final answer accuracy.
However, QA does not require correct attribution or inferential validity; a model may reach the correct answer via shortcut correlations or spurious visual cues.

\begin{table}[t]
\centering
\renewcommand{\arraystretch}{1.1}
\begin{adjustbox}{max width=\linewidth}
\begin{tabular}{lccc}
\toprule
\textbf{Metric} 
& \textbf{Zero-shot} 
& \textbf{Standard ICL} 
& \textbf{VLCG (Best)} \\
\midrule
\textbf{Causal Attribution (CA)} & 0.458 & 0.455 & \textbf{0.488} \\
\textbf{Causal Inference (CI)}   & 0.652 & 0.654 & \textbf{0.690} \\
\textbf{VQA Accuracy}            & 0.763 & 0.763 & \textbf{0.768} \\
\midrule
BLEU (reasoning)                 & 0.164 & 0.163 & \textbf{0.177} \\
ROUGE (reasoning)                & 0.266 & 0.264 & \textbf{0.273} \\
\bottomrule
\end{tabular}
\end{adjustbox}
\caption{
Performance comparison between zero-shot prompting, standard in-context learning (ICL), and VLCG-augmented prompting (best configuration). CA and CI measure relevance identification and inferential consistency, while BLEU/ROUGE assess surface-level reasoning overlap.
}
\vspace{-1.3em}
\label{tab:result_table}
\end{table}

\section{Results}
\subsection{Overall Performance}
Table~\ref{tab:result_table} reports performance across three prompting settings.
Standard in-context learning yields negligible gains over zero-shot prompting. CA slightly decreases (0.458 $\rightarrow$ 0.455), while CI and QA remain nearly unchanged. This indicates that simply providing question–answer exemplars does not meaningfully improve causal relevance identification. In some cases, ICL may even introduce noise by encouraging surface pattern imitation rather than structured reasoning.
In contrast, VLCG-augmented prompting improves both Causal Attribution and Causal Inference. CA increases from 0.458 to 0.488 (+6.6\% relative improvement),
while CI improves from 0.652 to 0.690 (+5.8\% relative improvement).
Notably, the improvement in CI is larger in absolute magnitude than CA, suggesting that structured graphs not only help identify relevant attributes, but also stabilize their composition into coherent reasoning chains.

Despite gains in CA and CI, VQA accuracy remains nearly constant (0.763 $\rightarrow$ 0.768).
This decoupling indicates that correct answers can be obtained even when relevance identification is imperfect.
In other words, models may exploit shortcut visual correlations to obtain correct answers,
while failing to construct stable causal explanations.
These results empirically support our claim that accuracy alone is insufficient to diagnose visual causal reasoning.
BLEU and ROUGE show modest improvements under VLCG prompting.
However, the relative gains in lexical overlap are smaller than those observed in CA and CI.
This suggests that improvements stem primarily from structured relevance alignment rather than superficial textual similarity.
Overall, the findings indicate that explicit causal structure acts as a relevance prior: it constrains the model’s attention to causally grounded attributes and reduces reliance on spurious cues. Importantly, this improvement manifests in attribution and inferential consistency, even when final answer accuracy remains stable.

\subsection{Qualitative Analysis}
\begin{figure}[!h]
  \includegraphics[width=0.98\columnwidth]{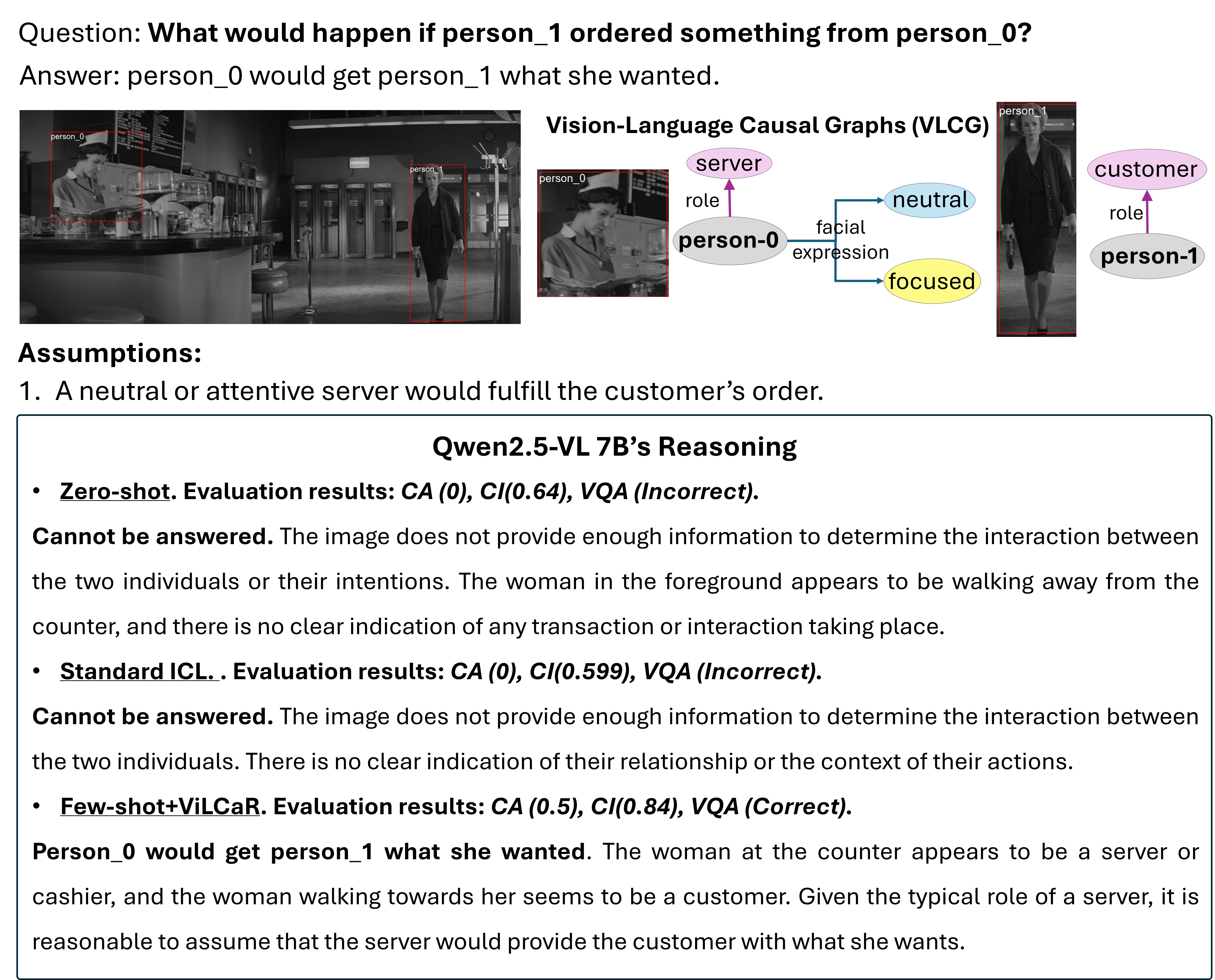}
  \caption{Reasonings from Qwen2.5-VL 7B model with: (1) Zero-shot, (2) Standard ICL, and (3) VLCG-augmented. Compared to the baselines, the model with ViLCaR injection is able to identify information about the role of the people relevant to the question.}
  \label{fig:qualitative_analysis_case_1}
  \vspace{-1em}
\end{figure}

Figure~\ref{fig:qualitative_analysis_case_1} presents a representative example illustrating how structured causal guidance alters model behavior under identical visual input.
In both zero-shot and standard ICL settings, the model concludes that the question cannot be answered.
Although the visual scene clearly contains two individuals positioned at a counter,
the model fails to identify their functional roles (e.g., \emph{server} and \emph{customer}).
Consequently, Causal Attribution (CA) is 0, indicating complete failure to retrieve relevant causal attributes.
Without role identification, the model cannot compose a valid reasoning chain, leading to incorrect QA predictions.
Notably, the failure is not due to perceptual inability but to relevance selection: the model describes visible entities but does not ground them in task-relevant causal roles.
When provided with structured VLCGs, the model explicitly identifies role attributes and links them to the assumption:
\emph{“A neutral or attentive server would fulfill the customer’s order.”}
This enables the model to construct a coherent causal chain from role identification to outcome prediction.
Hence, CA increases from 0 to 0.5, CI rises from 0.64/0.599 to 0.84, and the final QA prediction becomes correct. Importantly, the visual evidence remains unchanged across settings; the behavioral difference arises solely from structured relevance guidance.

This example highlights a key distinction in our framework: zero-shot and ICL failures stem from under-specification of causal roles, whereas VLCG prompting reduces ambiguity by explicitly constraining the reasoning space.
Rather than adding new perceptual information,
the graph acts as a relevance prior that encourages the model to ground its reasoning in causally meaningful variables. This supports our central claim that structured causal guidance improves attribution and inferential consistency, even when raw visual input remains constant.

\section{Conclusion}
We introduced ViLCaR, a benchmark for diagnosing failure modes in visual causal reasoning for LVLMs through structured VLCGs. 
By grounding causal attributes and assumptions in visual scenes, VLCGs provide a relevance prior for controlled analysis of reasoning behavior.
Our results show that answer accuracy alone is insufficient to diagnose reasoning quality, and that structured causal guidance improves attribution and inferential consistency even when accuracy remains unchanged.
Although our metrics rely on approximate semantic alignment, they provide a scalable diagnostic framework for analyzing reasoning failures, and future work may refine them through LLM-as-a-judge or human-in-the-loop validation.
Overall, ViLCaR offers a principled testbed for studying how LVLMs select, compose, and ground causally relevant information in visual reasoning tasks.

\bibliographystyle{ACM-Reference-Format}
\bibliography{sample-base}

\end{document}